\documentclass[%
 reprint,
 amsmath,amssymb,
 aps, 
 pre,
 showkeys
]{revtex4-1}

\usepackage{graphicx}% Include figure files
\usepackage{dcolumn}% Align table columns on decimal point
\usepackage{bm}% bold math
\usepackage{algpseudocode, algorithm}

\usepackage{my-style}
\graphicspath{{figs/}}

\allowdisplaybreaks[4]

\begin{document}

\preprint{APS/123-QED}

\title{Solving Non-parametric Inverse Problem in Continuous Markov Random Field using Loopy Belief Propagation}% Force line breaks with \\

\author{Muneki Yasuda}  
\affiliation{%
 Graduate School of Science and Engineering, Yamagata University.
}%
\author{Shun Kataoka}  
\affiliation{%
 Graduate School of Information Sciences, Tohoku University.
}%


\begin{abstract}
In this paper, we address the inverse problem, or the statistical machine learning problem, in Markov random fields with a non-parametric pair-wise energy function with continuous variables. 
The inverse problem is formulated by maximum likelihood estimation. 
The exact treatment of maximum likelihood estimation is intractable because of two problems: 
(1) it includes the evaluation of the partition function and 
(2) it is formulated in the form of functional optimization.  
We avoid Problem (1) by using Bethe approximation. Bethe approximation is an approximation technique equivalent to the loopy belief propagation. 
Problem (2) can be solved by using orthonormal function expansion. 
Orthonormal function expansion can reduce a functional optimization problem to a function optimization problem. 
Our method can provide an analytic form of the solution of the inverse problem within the framework of Bethe approximation.
\end{abstract}

\pacs{Valid PACS appear here}% PACS, the Physics and Astronomy
%Use showkeys class option if keyword
\maketitle

%\tableofcontents

\section{Introduction} \label{sec:Introduction}

Boltzmann machine learning, which is known as the inverse Ising problem in statistical mechanics, 
is one of the important problems in the statistical machine learning field and has a long history. 
Suppose that we have sample points, i.e., data points, stochastically generated from an unknown distribution (referred to as a generative model). 
The task of statistical machine learning is to specify the unknown distribution using only the sample points. 
In standard Boltzmann machine learning, we assume that the generative model that generates data points is an Ising model, 
and prepare an Ising model (referred to as the learning model) with controllable parameters, e.g., external fields and exchange interactions. 
The Boltzmann machine learning is achieved by optimizing the values of the controllable parameters in the learning model through maximum likelihood estimation. 

Unfortunately, we cannot perform Boltzmann machine learning exactly because of the computational cost. 
Therefore, many approximations for Boltzmann machine learning have been proposed. 
In particular, approximations based on mean-field methods have been developed in the field of statistical mechanics~\cite{Roudi2009}: 
mean-field approximation~\cite{Kappen1998}, Bethe approximation~\cite{Parise2005,Yasuda2006,Yasuda2009,Mezard2009,Marinari2010,Federico2012,Berg2012,Cyril2013}, Plefka expansion~\cite{TTanaka1998,Monasson2009}, and so on.  
In many of these methods, we can obtain the solution to the maximum likelihood estimation analytically. 
However, they are applicable to only an Ising-type learning model, that is, the variables in the model are binary and the energy function of the model is a quadratic form of the variables.

We proposed a method for a more general situation that uses Bethe approximation and orthonormal function expansion~\cite{Yasuda2012}. 
Using the method, we can solve the inverse problem with general pair-wise Markov random fields and obtain the solution analytically. 
However, this method cannot be applied to Markov random fields with continuous variables.

In this paper, we propose a method for solving the inverse problem in general pair-wise Markov random fields with continuous variables, which is an extension of our previous method~\cite{Yasuda2012}. 
The proposed method can give us the analytical solution of the inverse problem. 
This is the main contribution of this paper. 
In this paper, we refer to a pair-wise Markov random field with continuous variables as a continuous Markov random field (CMRF).

The remainder of this paper is organized as follows. 
In Sec. \ref{sec:LBP_for_CMRF}, we explain loopy belief propagation (LBP) in a CMRF. 
LBP is equivalent to Bethe approximation~\cite{GLBP2005,CVM-review2005}.
We formulate the inverse problem in a CMRF in Sec. \ref{sec:InverseProblem}, as well as its Bethe approximation. 
Our method is shown in Sec. \ref{sec:ProposedMethod}. 
In this section, we derive the solution to the inverse problem using the Bethe approximation shown in Sec. \ref{sec:InverseProblem}. 
Since the solution is obtained in the form of infinite series, it cannot be implemented as it is. 
We describe a means of implementing our method and show the results of numerical experiments in Sec. \ref{sec:implementation}.
We conclude the paper with some remarks in Sec. \ref{sec:conclusion}.

\section{Formalism of Loopy Belief Propagation in Continuous Markov Random Field}
\label{sec:LBP_for_CMRF}

Consider an undirected graph $G(V, E)$, where $V =\{ 1,2,\ldots, n\}$ is the set of nodes and $E$ is the set of undirected links. 
We denote the link between nodes $i$ and $j$ by $\{i,j\}$. 
Because the links have no direction, $\{i,j\}$ and $\{j,i\}$ indicate the same link.
On the undirected graph, we define the non-parametrized pair-wise energy function as 
\begin{align}
\Psi(\bm{x}) := -\sum_{i \in V}\theta_i(x_i) - \sum_{\{i,j\} \in E}w_{\{i,j\}}(\bm{x}_{\{i,j\}}),
\label{eqh:Hamiltonian}
\end{align}
where $\theta_i(x_i)$ is the energy on node $i$, $w_{\{i,j\}}(\bm{x}_{\{i,j\}})$ is the energy on link $\{i,j\}$, and $\bm{x}_{\{i,j\}} =\bm{x}_{\{j,i\}}= \{x_i ,x_j\}$. 
We regard $w_{\{i,j\}}(\bm{x}_{\{i,j\}})$ as the same function as $w_{\{j,i\}}(\bm{x}_{\{i,j\}})$.
With the energy function, we define the CMRF as 
\begin{align}
P(\bm{x}) := \frac{1}{Z} \exp\big(-\Psi(\bm{x})\big),
\label{eqn:CMRF}
\end{align}
where $\bm{x} =\{x_i \in \mcal{X} \mid i \in V\}$ represents the continuous random variables over the continuous space $\mcal{X}$, 
and $Z$ is the partition function defined as
\begin{align*}
Z:=\int_{\mcal{X}^n}\exp\big(-\Psi(\bm{x})\big)d\bm{x},
\end{align*}
where $\int_{\mcal{X}^n}f(\bm{x})d\bm{x}$ denotes the multiple integration over whole variables, $\int_{\mcal{X}} \cdots \int_{\mcal{X}} f(\bm{x})dx_1 \cdots dx_{n}$, 
and $\int_{\mcal{X}}$ denotes the integral over $\mcal{X}$.
$\theta_i(x_i)$ and $w_{\{i,j\}}(\bm{x}_{\{i,j\}})$ are arbitrary functions of the assigned variables. 

Given the CMRF, it is difficult to evaluate its marginal distributions because of  the existence of intractable multiple integration. 
LBP is one of the most effective methods for approximately evaluating marginal distributions 
and is the same as Bethe approximation in statistical mechanics. 
LBP can be obtained from the minimum condition of the variational Bethe free energy of the CMRF in Eq. (\ref{eqn:CMRF}). 
We denote the marginal distribution over $x_i$ by $b_i(x_i)$ and that over $x_i$ and $x_j$, which are neighboring pair of nodes, 
by $\xi_{\{i,j\}}(\bm{x}_{\{i,j\}})$. 
These marginal distributions are sometimes called \textit{beliefs} in the context of LBP. 
We regard $\xi_{\{j,i\}}(\bm{x}_{\{i,j\}})$ as the same belief as $\xi_{\{i,j\}}(\bm{x}_{\{i,j\}})$.
In the context of the cluster variation method~\cite{CVM1951,GLBP2005}, the variational Bethe free energy of the CMRF is expressed as
\begin{widetext}
\begin{align}
\mcal{F}[\bm{b},\bm{\xi}]
&:= \int_{\mcal{X}^n} \Psi(\bm{x}) P(\bm{x}) d\bm{x}+\sum_{i \in V}(1 - |\partial_i|)\int_{\mcal{X}}b_i(x_i)\ln b_i(x_i) dx_i 
+\sum_{\{i,j\} \in E}\int_{\mcal{X}^2}\xi_{\{i,j\}}(\bm{x}_{\{i,j\}})\ln \xi_{\{i,j\}}(\bm{x}_{\{i,j\}})dx_i dx_j\nn
&\>=-\sum_{i \in V}\int_{\mcal{X}}\theta_i(x_i)b_i(x_i) dx_i 
-\sum_{\{i,j\} \in E}\int_{\mcal{X}^2}w_{\{i,j\}}(\bm{x}_{\{i,j\}})\xi_{\{i,j\}}(\bm{x}_{\{i,j\}})dx_i dx_j 
+\sum_{i \in V}(1 - |\partial_i|)\int_{\mcal{X}}b_i(x_i)\ln b_i(x_i) dx_i \nn
\aleq\>
+\sum_{\{i,j\} \in E}\int_{\mcal{X}^2}\xi_{\{i,j\}}(\bm{x}_{\{i,j\}})\ln \xi_{\{i,j\}}(\bm{x}_{\{i,j\}})dx_i dx_j,
\label{eqn:VariatoinalBetheFreeEnergy}
\end{align}
\end{widetext}
where $\partial_i = \{j \mid \{i,j\} \in E\}$ is the set of nodes connected to node $i$. 
The variational Bethe free energy is regarded as the functional with respect to $\bm{b} = \{b_i (x_i) \mid i \in V\}$ and $\bm{\xi} = \{\xi_{\{i,j\}}(\bm{x}_{\{i,j\}}) \mid \{i,j\} \in E\}$. 
The beliefs, that minimize the variational Bethe free energy, are regarded as the Bethe approximation of the corresponding marginal distributions. 
From the extremal condition of the variational Bethe free energy under the normalizing constraints, 
\begin{align}
\int_{\mcal{X}}b_i(x_i) dx_i = \int_{\mcal{X}^2}\xi_{\{i,j\}}(\bm{x}_{\{i,j\}}) dx_idx_j = 1,
\label{eqn:NormalizingConstraint}
\end{align} 
and the marginalizing constraints, 
\begin{align}
\int_{\mcal{X}}\xi_{\{i,j\}}(\bm{x}_{\{i,j\}}) dx_i &= b_j(x_j),
\label{eqn:MarginalizingConstraint-i}\\
\int_{\mcal{X}}\xi_{\{i,j\}}(\bm{x}_{\{i,j\}}) dx_j &= b_i(x_i),
\label{eqn:MarginalizingConstraint-j}
\end{align}
we obtain the message-passing equation (MPE) 
\begin{align}
m_{i\to j}(x_j) =  \frac{1}{Z_{i \to j}}\int_{\mcal{X}} \pi_{i\setminus j}(x_i) e^{w_{\{i,j\}}(\bm{x}_{\{i,j\}})} dx_i, 
\label{eqn:MessagePassingEquation}
\end{align}
where the constant $Z_{i \to j}$ is frequently set to 
\begin{align}
Z_{i \to j} := \int_{\mcal{X}^2}\pi_{i\setminus j}(x_i) e^{w_{\{i,j\}}(\bm{x}_{\{i,j\}})} dx_i dx_j
\label{eqn:Def-Z_{i->j}}
\end{align}
to normalize the messages.
The distribution $\pi_{i \setminus j}(x_i)$ is defined as
\begin{align}
\pi_{i \setminus j}(x_i):= \frac{e^{\theta_i(x_i)}\prod_{k \in \partial_i \setminus \{j\}}m_{k \to i}(x_i)}
{\int_{\mcal{X}} e^{\theta_i(x_i)}\prod_{k \in \partial_i \setminus \{j\}}m_{k \to i}(x_i) dx_i}.
\label{eqn:Def-pi}
\end{align}
The quantity $m_{i \to j}(x_j)$ is the normalized message (or the effective field) from node $i$ to node $j$, which is non-negative and originates from the Lagrange multipliers 
appearing in the conditional minimization of the variational Bethe free energy. 
The two different messages, $m_{i \to j}(x_j)$ and $m_{j \to i}(x_i)$, are defined on link $\{i,j\}$.
The beliefs (the approximate marginal distributions) are computed from the messages as
\begin{align}
b_i(x_i) &\propto  e^{\theta_i(x_i)} \prod_{k \in \partial_i }m_{k \to i}(x_i), 
\label{eqn:belief-1}\\
\xi_{\{i,j\}}(\bm{x}_{\{i,j\}}) &\propto e^{w_{\{i,j\}}(\bm{x}_{\{i,j\}})} \pi_{i \setminus j}(x_i)\pi_{j \setminus i}(x_j).
\label{eqn:belief-2}
\end{align}

In principle, by solving the MPE in Eq. (\ref{eqn:MessagePassingEquation}), we can compute the one- and two-variable marginal distributions using Eqs. (\ref{eqn:belief-1}) and (\ref{eqn:belief-2}). 
However, finding the functional forms of the messages is not straightforward, because the messages are continuous functions over $\mcal{X}$, 
and therefore, the MPE we have to solve is an integral equation.
Some methods that are based mainly on a stochastic method have been developed for approximately solving the MPE~\cite{NonParaBP2003, PartBP2009, BPforCMRF2013}.

\section{Inverse Problem in Continuous Markov Random Field}
\label{sec:InverseProblem}

In this section, we consider the inverse problem, in other words, the machine learning problem, for the CMRF in Eq. (\ref{eqn:CMRF}). 
The inverse problem for the CMRF can be solved by maximum likelihood estimation.
Given $N$ data points $\mcal{D} = \{\mathbf{x}^{(\mu)} \in \mcal{X}^{n} \mid i = 1,2,\ldots, N\}$, we define the log-likelihood functional as
\begin{align}
l[\bm{\theta}, \bm{w}]:= \frac{1}{N} \sum_{\mu = 1}^N \ln P(\mathbf{x}^{(\mu)}),
\label{eqn:log-likelihood} 
\end{align}
where $\bm{\theta}$ and $\bm{w}$ are the set of functions $\theta_i(x_i)$ and $w_{\{i,j\}}(\bm{x}_{\{i,j\}})$ respectively in the exponent in Eq. (\ref{eqn:CMRF}). 
The goal of the maximum likelihood estimation is to find the functions $\bm{\theta}$ and $\bm{w}$ that maximize the log-likelihood functional.
Eq. (\ref{eqn:log-likelihood}) can be rewritten as
\begin{align}
l[\bm{\theta}, \bm{w}] &= -\frac{1}{N} \sum_{\mu = 1}^N\Psi(\mathbf{x}^{(\mu)})-\ln Z.
\label{eqn:log-likelihood-trans} 
\end{align}
However, the maximization problem of the log-likelihood functional is intractable because of the existence of the partition function.

To avoid evaluating the intractable partition function, 
we approximate the log-likelihood functional using LBP, i.e., Bethe approximation. 
The Bethe approximation of the log-likelihood functional in Eq. (\ref{eqn:log-likelihood-trans}) can be expressed 
by using the variational Bethe free energy shown in Eq. (\ref{eqn:VariatoinalBetheFreeEnergy}) as
\begin{align}
l_{\mrm{Bethe}}[\bm{\theta}, \bm{w}] 
:= - \frac{1}{N} \sum_{\mu = 1}^N\Psi(\mathbf{x}^{(\mu)}) + \min_{\bm{b},\bm{\xi}} \mcal{F}[\bm{b},\bm{\xi}].
\label{eqn:log-likelihood_BetheApprox}
\end{align}
We refer to this as the Bethe log-likelihood functional. 
The main purpose of this study was to maximize the Bethe log-likelihood functional with respect to the functions $\bm{\theta}$ and $\bm{w}$.
The solution obtained by maximizing Eq. (\ref{eqn:log-likelihood_BetheApprox}), of course coincides to that obtained by the true maximum likelihood estimation 
when the CMRF has a tree structure, because Bethe approximation is exact in tree systems. 
However, the maximization of the Bethe log-likelihood functional is not straightforward for the following reasons. 
The variations of the functional with respect to $\bm{\theta}$ and $\bm{w}$ are
\begin{align*}
\frac{\delta l_{\mrm{Bethe}}[\bm{\theta}, \bm{w}]}{\delta \theta_i(x_i)}
&=\frac{1}{N} \sum_{\mu = 1}^N \delta(x_i - \mrm{x}_i^{(\mu)}) - b_i(x_i), \\
\frac{\delta l_{\mrm{Bethe}}[\bm{\theta}, \bm{w}]}{\delta w_{\{i,j\}}(\bm{x}_{\{i,j\}})}
&=\frac{1}{N} \sum_{\mu = 1}^N \delta(x_i - \mrm{x}_i^{(\mu)}) \delta(x_j - \mrm{x}_j^{(\mu)})\nn
\aleq - \xi_{\{i,j\}}(\bm{x}_{\{i,j\}}),
\end{align*}
where $b_i(x_i)$ and $\xi_{\{i,j\}}(\bm{x}_{\{i,j\}})$ are the beliefs minimizing the variational Bethe free energy, in other words, 
the solution to the LBP presented in the previous section. 
This variation means that we have to find $\bm{\theta}$ and $\bm{w}$ that satisfy the relations 
\begin{align}
\frac{1}{N} \sum_{\mu = 1}^N f(\mrm{x}_i^{(\mu)})  
=\int_{\mcal{X}} f(x_i) b_i(x_i)dx_i
\label{eqn:BetheMLE_theta}
\end{align}
and 
\begin{align}
\frac{1}{N} \sum_{\mu = 1}^N g(\mrm{x}_i^{(\mu)}, \mrm{x}_j^{(\mu)}) 
=\int_{\mcal{X}} g(x_i, x_j)  \xi_{\{i,j\}}(\bm{x}_{\{i,j\}})dx_idx_j
\label{eqn:BetheMLE_w}
\end{align}
for any test functions $f(x_i)$ and $g(x_i, x_j)$. 
Thus, if we could obtain the solution of the LBP, by using a method that has already proposed~\cite{NonParaBP2003, PartBP2009, BPforCMRF2013}, 
the solution to the maximization of the Bethe log-likelihood functional is not immediately obtained.

\section{Proposed Method}
\label{sec:ProposedMethod}

In this section, we propose a method to solve the maximization problem of the Bethe log-likelihood function in Eq. (\ref{eqn:log-likelihood_BetheApprox}) in terms of orthonormal function expansion. 
Via orthonormal function expansion, we can reduce the functional maximization problem in the previous section 
to a tractable function maximization problem. 
The basic idea of our method is similar to that presented in our previous paper~\cite{Yasuda2012}.

\subsection{Orthonormal Function System}
\label{sec:OrthonormalFunctionSystem}

Before deriving our method, we introduce an orthonormal function system $\{ \phi_s(x) \mid s = 0, 1,2,\ldots\}$ over $\mcal{X}$   
satisfying 
\begin{align}
\int_{\mcal{X}}  \phi_s(x)\phi_t(x) dx = \delta_{s,t},
\label{eqn:orthonormalRelation}
\end{align}
where $\delta_{s,t}$ is the Kronecker delta function.
By using the orthonormal function system, function $f(x)$ over $\mcal{X}$ is expanded as 
\begin{align}
f(x) = \sum_{s = 0}^{\infty} \alpha_s \phi_s(x), 
\label{eqn:orthonormalFunctionsExpansion}
\end{align}
where the expanding coefficients are given by 
\begin{align}
\alpha_s = \int_{\mcal{X}} f(x) \phi_s(x) dx.
\label{eqn:orthonormalFunctionsExpansion-Coeff}
\end{align}
The orthonormal function expansion in Eq. (\ref{eqn:orthonormalFunctionsExpansion}) plays an important role in our method. 

In the following, we assume that $\mcal{X}$ is the finite space, $\mcal{X} = [\alpha, \beta]$,  
and that $\phi_0(x)$ is constant over $\mcal{X}$, i.e., 
\begin{align}
\phi_0(x) = \frac{1}{\sqrt{ \chi}},\quad \chi := \beta - \alpha.
\label{eqn:phi0}
\end{align}
From Eqs. (\ref{eqn:orthonormalRelation}) and (\ref{eqn:phi0}), we have 
\begin{align}
\int_{\mcal{X}} \phi_s(x) dx =
\begin{cases}
\sqrt{\chi} & s = 0\\
0 & s > 0
\end{cases}
.
\label{eqn:orthonormalSingleRelation}
\end{align}
Examples of this orthonormal function are described in Appendix \ref{sec:app:Ex_OrthonormalFunctionSystem}. 
We use Eqs. (\ref{eqn:orthonormalRelation}), (\ref{eqn:phi0}), and (\ref{eqn:orthonormalSingleRelation}) frequently throughout the paper.

The orthonormal function expansion introduced in this section plays a central role in our proposed method described in the following. 
However, a similar idea can be useful for solving the LBP in Sec. \ref{sec:LBP_for_CMRF}. 
Indeed, a method for solving the LBP was proposed by using orthonormal function expansion~\cite{BPforCMRF2013}.

\subsection{Variational Bethe Free Energy with Orthonormal Function Expansion}

First, we rewrite the CMRF in Eq. (\ref{eqn:CMRF}) by expanding $\bm{\theta}$ and $\bm{w}$. 
By using the orthonormal function expansion in Eq. (\ref{eqn:orthonormalFunctionsExpansion}), 
the functions $\bm{\theta}$ and $\bm{w}$ can be expanded as follows.
\begin{align}
\theta_i(x_i) &= \sum_{s = 0}^{\infty}h_i^{(s)} \phi_s(x_i) = \sum_{s = 1}^{\infty}h_i^{(s)} \phi_s(x_i) + \mrm{constant}
\label{eqn:theta_expand}
\end{align}
and
\begin{align}
w_{\{i,j\}}(\bm{x}_{\{i,j\}}) &= \sum_{s,t = 0}^{\infty}J_{\{i,j\}}^{(s,t)} \phi_s(x_i)\phi_t(x_j)\nn
&= \frac{1}{\sqrt{\chi}}\sum_{s= 1}^{\infty}\big( J_{\{i,j\}}^{(s,0)}\phi_s(x_i) + J_{\{i,j\}}^{(0,s)}\phi_s(x_j)\big)\nn
\aleq + \sum_{s,t = 1}^{\infty}J_{\{i,j\}}^{(s,t)} \phi_s(x_i)\phi_t(x_j) + \mrm{constant},
\label{eqn:wij_expand}
\end{align}
where, from Eq. (\ref{eqn:orthonormalFunctionsExpansion-Coeff}), the expanding coefficients are  
\begin{align}
h_i^{(s)}&:=\int_{\mcal{X}} \theta_i(x_i) \phi_s(x_i) dx_i, 
\label{eqn:def-coeff-hi}\\
J_{\{i,j\}}^{(s,t)} &:=\int_{\mcal{X}^2} w_{\{i,j\}}(\bm{x}_{\{i,j\}}) \phi_s(x_i)\phi_t(x_j) dx_i dx_j.
\label{eqn:def-coeff-Jij}
\end{align}
It is noteworthy that, from the symmetric property of $w_{\{i,j\}}(\bm{x}_{\{i,j\}})$, $J_{\{i,j\}}^{(s,t)} =J_{\{j,i\}}^{(t,s)}$ is satisfied. 
In Eqs. (\ref{eqn:theta_expand}) and (\ref{eqn:wij_expand}), 
Eq. (\ref{eqn:phi0}) is used.
Using Eqs. (\ref{eqn:theta_expand}) and (\ref{eqn:wij_expand}), we can rewrite the energy function in Eq. (\ref{eqh:Hamiltonian}) as
\begin{align}
\Psi(\bm{x}) = \Psi^{\dagger}(\bm{x}; \bm{H}, \bm{J}) + C_0,
\label{eqn:Hamiltonian_trans}
\end{align}
where
\begin{align}
\Psi^{\dagger}(\bm{x}; \bm{H}, \bm{J})&:=-\sum_{i \in V}\sum_{s = 1}^{\infty}H_i^{(s)}  \phi_s(x_i) \nn
\aleq \>
- \sum_{\{i,j\}\in E} \sum_{s,t = 1}^{\infty} J_{\{i,j\}}^{(s,t)} \phi_s(x_i)\phi_t(x_j)
\label{eqn:Hamiltonian_expand}
\end{align}
and
\begin{align*}
H_i^{(s)} := h_i^{(s)} + \frac{1}{\sqrt{\chi}}\sum_{j \in \partial_i}J_{\{i,j\}}^{(s,0)}.
\end{align*} 
The constant $C_0$ in Eq. (\ref{eqn:Hamiltonian_trans}) originates from the constants in Eqs. (\ref{eqn:theta_expand}) and (\ref{eqn:wij_expand}).
Therefore, using the new energy function, the CMRF in Eq. (\ref{eqn:CMRF}) can be rewritten as
\begin{align}
P(\bm{x})= P(\bm{x}\mid  \bm{H}, \bm{J}) \propto \exp\big( - \Psi^{\dagger}(\bm{x}; \bm{H}, \bm{J})\big).
\label{eqn:CMRF_expand}
\end{align}
This rewriting makes the CMRF the parametric model, parameterized by 
$\bm{H}=\{H_i^{(s)} \mid i \in V , s \geq 1\}$ and $\bm{J}=\{J_{\{i,j\}}^{(s,t)} \mid \{i,j\} \in E , s \geq 1 ,t \geq 1\}$. 
In Eq. (\ref{eqn:CMRF_expand}), the constant in Eq. (\ref{eqn:Hamiltonian_trans}) is neglected, 
because it is irrelevant to the distribution.

Now, we introduce the orthonormal function expansions of the beliefs in the variational Bethe free energy, as follows.
\begin{align}
b_i(x_i) &= \sum_{s = 0}^{\infty}c_i^{(s)} \phi_s(x_i), 
\label{eqn:belief-1_expand}\\
\hat{\xi}_{\{i,j\}}(\bm{x}_{\{i,j\}}) &= \sum_{s,t = 0}^{\infty}d_{\{i,j\}}^{(s,t)} \phi_s(x_i)\phi_t(x_j). 
\label{eqn:belief-2_expand}
\end{align} 
From Eq. (\ref{eqn:orthonormalFunctionsExpansion-Coeff}), the expanding coefficients are 
\begin{align}
c_i^{(s)}&:=\int_{\mcal{X}} b_i(x_i) \phi_s(x_i) dx_i, 
\label{eqn:def-ci^s}\\
d_{\{i,j\}}^{(s,t)} &:=\int_{\mcal{X}^2} \xi_{\{i,j\}}(\bm{x}_{\{i,j\}}) \phi_s(x_i)\phi_t(x_j) dx_i dx_j.
\label{eqn:def-dij^st}
\end{align}
The beliefs must satisfy the normalizing constraints in Eq. (\ref{eqn:NormalizingConstraint})
and the marginalizing constraints in Eqs. (\ref{eqn:MarginalizingConstraint-i}) and (\ref{eqn:MarginalizingConstraint-j}).
From Eqs. (\ref{eqn:NormalizingConstraint}), (\ref{eqn:phi0}), (\ref{eqn:def-ci^s}), and (\ref{eqn:def-dij^st}), we have
\begin{align*}
c_i^{(0)} = \frac{1}{\sqrt{\chi}}, \quad d_{\{i,j\}}^{(0,0)} = \frac{1}{\chi}.
\end{align*} 
From Eqs. (\ref{eqn:MarginalizingConstraint-i}), (\ref{eqn:MarginalizingConstraint-j}), (\ref{eqn:phi0}), (\ref{eqn:def-ci^s}), and (\ref{eqn:def-dij^st}), 
we obtain
\begin{align*}
d_{\{i,j\}}^{(s,0)} =\frac{c_i^{(s)}}{\sqrt{\chi}}, \quad d_{\{i,j\}}^{(0,s)} =\frac{c_j^{(s)}}{\sqrt{\chi}}.
\end{align*}
From the above equations, the beliefs in Eqs. (\ref{eqn:belief-1_expand}) and (\ref{eqn:belief-2_expand}) can be expressed as
\begin{align}
b_i(x_i) &= \frac{1}{\chi} + \sum_{s = 1}^{\infty}c_i^{(s)} \phi_s(x_i)=:b_i(x_i \mid \bm{c}), 
\label{eqn:belief-1_expand-trans}\\
\xi_{\{i,j\}}(\bm{x}_{\{i,j\}}) &= \frac{1}{\chi^2}  + \frac{1}{ \chi}\sum_{s = 1}^{\infty}\big(c_i^{(s)} \phi_s(x_i) \nn
\aleq
+ c_j^{(s)} \phi_s(x_j)\big)+ \sum_{s,t = 1}^{\infty}d_{\{i,j\}}^{(s,t)} \phi_s(x_i)\phi_t(x_j) \nn
&=: \xi_{\{i,j\}}(\bm{x}_{\{i,j\}}\mid \bm{c},\bm{d}),
\label{eqn:belief-2_expand-trans}
\end{align} 
where $\bm{c} = \{c_i^{(s)} \mid i \in V , s \geq 1\}$ and $\bm{d} = \{d_{\{i,j\}}^{(s,t)} \mid \{i,j\} \in E, s \geq 1 ,t \geq 1\}$.
By using Eq. (\ref{eqn:orthonormalSingleRelation}), 
one can confirm that the beliefs in Eqs. (\ref{eqn:belief-1_expand-trans}) and (\ref{eqn:belief-2_expand-trans}) satisfy the normalization constraints and the marginal constraints
for any $\bm{c}$ and $\bm{d}$.

From Eqs. (\ref{eqn:Hamiltonian_expand}), (\ref{eqn:belief-1_expand-trans}), and (\ref{eqn:belief-2_expand-trans}), 
in the same way as in Eq. (\ref{eqn:VariatoinalBetheFreeEnergy}),
we formulate the variational Bethe free energy for the CMRF in Eq. (\ref{eqn:CMRF_expand}) as
\begin{align}
\mcal{F}(\bm{c}, \bm{d})&=-\sum_{i \in V}\sum_{s = 1}^{\infty}H_i^{(s)} c_i^{(s)}
- \sum_{\{i,j\} \in E}\sum_{s,t = 1}^{\infty}J_{\{i,j\}}^{(s,t)}d_{\{i,j\}}^{(s,t)} \nn
&+\sum_{i \in V}(1 - |\partial_i|)\int_{\mcal{X}}b_i(x_i \mid \bm{c})
\ln b_i(x_i \mid \bm{c}) dx_i \nn
&+\sum_{\{i,j\} \in E}\int_{\mcal{X}^2}\xi_{\{i,j\}}(\bm{x}_{\{i,j\}}\mid \bm{c},\bm{d}) \nn
&\times \ln \xi_{\{i,j\}}(\bm{x}_{\{i,j\}}\mid \bm{c},\bm{d})dx_i dx_j.
\label{eqn:VariatoinalBetheFreeEnergy_expand}
\end{align}
For specific $\bm{\theta}$ and $\bm{w}$, this variational Bethe free energy is not the functional, but the function of 
$\bm{c}$ and $\bm{d}$. 
The variational Bethe free energy in Eq. (\ref{eqn:VariatoinalBetheFreeEnergy_expand}) coincides with that in Eq. (\ref{eqn:VariatoinalBetheFreeEnergy}), 
except for the irrelevant constant neglected in Eq. (\ref{eqn:CMRF_expand}), i.e., 
\begin{align}
\mcal{F}[\bm{b},\bm{\xi}] = \mcal{F}(\bm{c}, \bm{d}) + C_0.
\label{eqn:VariationalBetheFreeEnergy_trans}
\end{align}

As mentioned above, the beliefs in Eqs. (\ref{eqn:belief-1_expand-trans}) and (\ref{eqn:belief-2_expand-trans}) satisfy the normalization constraints and the marginal constraints 
for any $\bm{c}$ and $\bm{d}$, so that we can minimize $\mcal{F}(\bm{c}, \bm{d})$ with no constraint. 
At the minimum point of $\mcal{F}(\bm{c}, \bm{d})$, $\bm{c}$ and $\bm{d}$ satisfy
\begin{align}
H_i^{(s)} &= (1 - |\partial_i|)\int_{\mcal{X}}\phi_s(x_i)\ln b_i(x_i \mid \bm{c}) dx_i \nn
&+\frac{1}{\chi}\sum_{j \in \partial_i}\int_{\mcal{X}^2}\phi_s(x_i)\ln  \xi_{\{i,j\}}(\bm{x}_{\{i,j\}}\mid \bm{c},\bm{d})dx_i dx_j, 
\label{eqn:etremal_VariationalBetheFreeEnergy_c}\\
J_{\{i,j\}}^{(s,t)}&=\int_{\mcal{X}^2}\phi_s(x_i)\phi_t(x_j)\ln \xi_{\{i,j\}}(\bm{x}_{\{i,j\}}\mid \bm{c},\bm{d})dx_i dx_j.
\label{eqn:etremal_VariationalBetheFreeEnergy_d}
\end{align}
Eqs. (\ref{eqn:etremal_VariationalBetheFreeEnergy_c}) and (\ref{eqn:etremal_VariationalBetheFreeEnergy_d}) are derived from 
the extremal condition of Eq. (\ref{eqn:VariatoinalBetheFreeEnergy_expand}) with respect to $c_i^{(s)}$ and $d_{\{i,j\}}^{(s,t)}$, respectively. 
In the derivation of these equations, we used Eq. (\ref{eqn:orthonormalSingleRelation}).

\subsection{Maximization of the Bethe Log-likelihood Function}
\label{sec:solution_BetheMLE}

By using the new energy function in Eqs. (\ref{eqn:Hamiltonian_expand}) and the variational Bethe free energy in Eq. (\ref{eqn:VariatoinalBetheFreeEnergy_expand}), 
the Bethe log-likelihood functional in Eq. (\ref{eqn:log-likelihood_BetheApprox}) is represented as
\begin{align}
l_{\mrm{Bethe}}(\bm{H}, \bm{J})=- \frac{1}{N} \sum_{\mu = 1}^N\Psi^{\dagger}(\mathbf{x}^{(\mu)}; \bm{H}, \bm{J})
+ \min_{\bm{c}, \bm{d}}\mcal{F}(\bm{c}, \bm{d}).
\label{eqn:log-likelihood_BetheApprox_expand}
\end{align}
This is the function with respect to $\bm{H}$ and $\bm{J}$ and we refer to this function as the Bethe log-likelihood function. 
Thus, the functional optimization problem of the maximum likelihood estimation is reduced to the function optimization problem. 
The Bethe log-likelihood function is equivalent to the Bethe log-likelihood functional in Eq. (\ref{eqn:log-likelihood_BetheApprox}), because, 
from Eqs. (\ref{eqn:Hamiltonian_trans}) and (\ref{eqn:VariationalBetheFreeEnergy_trans}),
\begin{align*}
l_{\mrm{Bethe}}[\bm{\theta}, \bm{w}]
&=- \frac{1}{N} \sum_{\mu = 1}^N\big(\Psi^{\dagger}(\mathbf{x}^{(\mu)}; \bm{H}, \bm{J}) + C_0\big) \nn
\aleq 
+ \min_{\bm{c}, \bm{d}}\big( \mcal{F}(\bm{c}, \bm{d}) + C_0\big)\nn
&=l_{\mrm{Bethe}}(\bm{H}, \bm{J}).
\end{align*}
Therefore, the maximization of the the Bethe log-likelihood function with respect to $\bm{H}$ and $\bm{J}$ 
is equivalent to the maximization of the Bethe log-likelihood functional with respect to $\bm{\theta}$ and $\bm{w}$.
At the maximum point of the Bethe log-likelihood function, we have equations for the expanding coefficients in Eqs. (\ref{eqn:belief-1_expand-trans}) and (\ref{eqn:belief-2_expand-trans}) as 
\begin{align}
\hat{c}_i^{(s)} &= \ave{\phi_s(x_i)}_{\mcal{D}}, 
\label{eqn:MaximumCondition_c}\\
\hat{d}_{\{i,j\}}^{(s,t)} &=  \ave{\phi_s(x_i)\phi_t(x_j)}_{\mcal{D}},
\label{eqn:MaximumCondition_d}
\end{align}
where $\ave{f(\bm{x})}_{\mcal{D}} := N^{-1}\sum_{\mu = 1}^N f(\mathbf{x}^{(\mu)})$ is the sample average over data points $\mcal{D}$. 
Coefficients $\hat{c}_i^{(s)}$ and $\hat{d}_{\{i,j\}}^{(s,t)}$ are the solutions to the minimization of the variational Bethe free energy in Eq. (\ref{eqn:VariatoinalBetheFreeEnergy_expand}), 
that is, the solutions to Eqs. (\ref{eqn:etremal_VariationalBetheFreeEnergy_c}) and (\ref{eqn:etremal_VariationalBetheFreeEnergy_d}). 
In the following, we denote the beliefs, the coefficients of which are fixed by Eqs. (\ref{eqn:MaximumCondition_c}) 
and (\ref{eqn:MaximumCondition_d}), by $\hat{\bm{b}}$ and $\hat{\bm{\xi}}$, i.e., 
$\hat{b}_i(x_i) := b_i(x_i \mid \hat{\bm{c}})$ and $ \hat{\xi}_{\{i,j\}}(\bm{x}_{\{i,j\}}):= \xi_{\{i,j\}}(\bm{x}_{\{i,j\}} \mid \hat{\bm{c}}, \hat{\bm{d}})$.

By substituting Eqs. (\ref{eqn:MaximumCondition_c}) and (\ref{eqn:MaximumCondition_d}) into Eqs. (\ref{eqn:etremal_VariationalBetheFreeEnergy_c}) and (\ref{eqn:etremal_VariationalBetheFreeEnergy_d}), 
we can obtain the solution, $\hat{\bm{H}}$ and $\hat{\bm{J}}$, to the maximization of the Bethe log-likelihood function in Eq. (\ref{eqn:log-likelihood_BetheApprox_expand}), and then, 
identify the energy function $\Psi^{\dagger}(\bm{x} \mid \hat{\bm{H}}, \hat{\bm{J}})$.  
It should be noted that the solution obtained by our method satisfies Eqs. (\ref{eqn:BetheMLE_theta}) and (\ref{eqn:BetheMLE_w}), 
which is easily confirmed as follows. 
A test function $f(x)$ is expanded as in Eq. (\ref{eqn:orthonormalFunctionsExpansion}).
Therefore, the left side of Eq. (\ref{eqn:BetheMLE_theta}) is
\begin{align*}
\frac{1}{N}\sum_{\mu = 1}^{N} f(x_i)=\frac{\alpha_0}{\chi} + \sum_{s = 1}^{\infty}\alpha_s \ave{\phi_s(x_i)}_{\mcal{D}}.
\end{align*}
On the other side, the right hand side of Eq. (\ref{eqn:BetheMLE_theta}) is 
\begin{align*}
\int_{\mcal{X}} f(x_i) \hat{b}_i(x_i) dx_i &= \frac{\alpha_0}{\chi} 
+ \sum_{s = 1}^{\infty}\alpha_s \int_{\mcal{X}} \phi_s(x_i)\hat{b}_i(x_i) \nn
&=\frac{\alpha_0}{\chi} 
+ \sum_{s = 1}^{\infty}\alpha_s \hat{c}_i^{(s)}, 
\end{align*}
where we use Eq. (\ref{eqn:belief-1_expand-trans}). 
From these equations and Eq. (\ref{eqn:MaximumCondition_c}), 
the solution obtained by our method satisfying Eq. (\ref{eqn:BetheMLE_theta}) is confirmed.  
Similarly, we can verify the equality in Eq. (\ref{eqn:BetheMLE_w}).

By using the method described above, within the framework of Bethe approximation 
we can identify the functional form of the energy function through the use of the given $N$ data points, 
and then obtain the resulting CMRF as 
\begin{align}
\hat{P}(\bm{x}) \propto \exp\big( - \Psi^{\dagger}(\bm{x}; \hat{\bm{H}}, \hat{\bm{J}})\big).
\label{eqn:trained-CMRF}
\end{align}
Unfortunately, one cannot computationally treat the infinite series in Eqs. (\ref{eqn:etremal_VariationalBetheFreeEnergy_c}) and (\ref{eqn:etremal_VariationalBetheFreeEnergy_d}). 
Thus, in practice, we truncate the infinite series and approximate them by a finite series obtained by the truncation. 
The details of this approximation are described in Sec. \ref{sec:Approx_For_Implementation}.

The proposed method includes the integration procedures (cf. Eqs. (\ref{eqn:etremal_VariationalBetheFreeEnergy_c}) and (\ref{eqn:etremal_VariationalBetheFreeEnergy_d})).  
The following rewriting allows us to identify the functional form of $\Psi^{\dagger}(\bm{x} \mid \hat{\bm{H}}, \hat{\bm{J}})$ without the integration procedures.
We now consider the energy function defined by
\begin{align}
\Psi^{\ddagger}(\bm{x}):= - \sum_{i \in V}(1 - |\partial_i|)\ln \hat{b}_i(x_i) 
- \sum_{\{i,j\} \in E} \ln \hat{\xi}_{\{i,j\}}(\bm{x}_{\{i,j\}}).
\label{eqn:Hamiltonian_belief}
\end{align} 
This energy function satisfies the relation
\begin{align}
\Psi^{\ddagger}(\bm{x}) = \Psi^{\dagger}(\bm{x} \mid \hat{\bm{H}}, \hat{\bm{J}}) + C_1,
\label{eqn:Hamiltonian_trans2}
\end{align}
where $C_1$ is the constant unrelated to $\bm{x}$ (see Appendix \ref{sec:app:RelationHamiltonians}). 
From the relation and Eq. (\ref{eqn:trained-CMRF}), we obtain the CMRF determined by the Bethe approximation of the MLE as 
\begin{align}
\hat{P}(\bm{x}) \propto \exp\big( - \Psi^{\ddagger}(\bm{x})\big),
\label{eqn:CMRF_trained}
\end{align}
and obtain the energy function in the form of Eq. (\ref{eqn:Hamiltonian_belief}). 

\section{Implementation}
\label{sec:implementation}

\subsection{Approximation for Implementation}
\label{sec:Approx_For_Implementation}

The beliefs, $\hat{b}_i(x_i)$ and $\hat{\xi}_{ij}(x_i,x_j)$, are expressed by an infinite series, as shown in Eqs. (\ref{eqn:belief-1_expand-trans}) and (\ref{eqn:belief-2_expand-trans}). 
Because an infinite series is not implementable, we approximate them by the truncation up to a finite order:
\begin{align}
\hat{b}_i^{(K)}(x_i) := \frac{1}{\chi} + \sum_{s = 1}^{K}\hat{c}_i^{(s)} \phi_s(x_i)
\label{eqn:Kapprox_bi}
\end{align}
and
\begin{align}
\hat{\xi}_{\{i,j\}}^{(K)}(\bm{x}_{\{i,j\}}) &:= \frac{1}{\chi^2}  + \frac{1}{\chi}\sum_{s = 1}^{K}\big(\hat{c}_i^{(s)} \phi_s(x_i) \nn
\aleq \>
+ \hat{c}_j^{(s)} \phi_s(x_j)\big)+ \sum_{s,t = 1}^{K}\hat{d}_{\{i,j\}}^{(s,t)} \phi_s(x_i)\phi_t(x_j),
\label{eqn:Kapprox_xi}
\end{align}
where the positive integer $K$ controls the order of the approximation. 
In the limit of $K \to \infty$, $\hat{b}_i^{(K)}(x_i)$ and $\hat{\xi}_{\{i,j\}}^{(K)}(\bm{x}_{\{i,j\}})$ coincide with 
$\hat{b}_i(x_i)$ and $\hat{\xi}_{\{i,j\}}(\bm{x}_{\{i,j\}})$, respectively. 
The approximate beliefs in Eqs. (\ref{eqn:Kapprox_bi}) and (\ref{eqn:Kapprox_xi}) are normalized for any $K > 0$.

Because of the above truncating approximation, the non-negativity of the beliefs may not be retained. 
Thus, to preserve the positivity of the beliefs, we have to make a further approximation to them. 
For a small positive value $\varepsilon$, we define distributions
\begin{align}
 \tilde{b}_i^{(K)}(x_i):= \frac{\max(\varepsilon, \hat{b}_i^{(K)}(x_i))}
{\int_{\mcal{X}}\max(\varepsilon, \hat{b}_i^{(K)}(x_i))dx_i}
\label{eqn:Kapprox_bi_positiveApproximation}
\end{align}
and regard the cut-off distribution as the approximation of $\hat{b}_i^{(K)}(x_i)$. 
If $\hat{b}_i^{(K)}(x_i) \geq \varepsilon$ over $\mcal{X}$, $\tilde{b}_i^{(K)}(x_i) = \hat{b}_i^{(K)}(x_i)$.
In a similar manner, we approximate $\hat{\xi}_{\{i,j\}}^{(K)}(\bm{x}_{\{i,j\}})$ by
\begin{align}
\tilde{\xi}_{\{i,j\}}^{(K)}(\bm{x}_{\{i,j\}}):= \frac{\max(\varepsilon, \hat{\xi}_{i,j}^{(K)}(x_i,x_j))}
{\int_{\mcal{X}^2}\max(\varepsilon, \hat{\xi}_{i,j}^{(K)}(x_i,x_j))dx_idx_j}.
\label{eqn:Kapprox_xi_positiveApproximation}
\end{align}
By using Eqs. (\ref{eqn:Kapprox_bi_positiveApproximation}) and (\ref{eqn:Kapprox_xi_positiveApproximation}) instead of $\hat{b}_i(x_i)$ and $\hat{\xi}_{ij}(x_i,x_j)$, 
the CMRF in Eq. (\ref{eqn:CMRF_trained}) is approximated by 
\begin{align}
\hat{P}(\bm{x})\approx \tilde{P}_K(\bm{x}) \propto \exp\big( - \tilde{\Psi}_K^{\ddagger}(\bm{x})\big),
\label{eqn:CMRF_trained_approx}
\end{align}
where
\begin{align}
&\tilde{\Psi}_K^{\ddagger}(\bm{x})\nn
&:= - \sum_{i \in V}(1 - |\partial_i|)\ln \tilde{b}_i^{(K)}(x_i)
- \sum_{\{i,j\} \in E} \ln  \tilde{\xi}_{\{i,j\}}^{(K)}(\bm{x}_{\{i,j\}})\nn
&= - \sum_{i \in V}(1 - |\partial_i|)\ln \big(\max(\varepsilon, \hat{b}_i^{(K)}(x_i))\big)\nn
\aleq
- \sum_{\{i,j\} \in E} \ln  \big(\max(\varepsilon, \hat{\xi}_{\{i,j\}}^{(K)}(\bm{x}_{\{i,j\}}))\big) + C_2
\label{eqn:Hamiltonian_belief_approx}
\end{align} 
is the approximation of Eq. (\ref{eqn:Hamiltonian_belief}).
Constant $C_2$, which originates from the denominators of Eqs. (\ref{eqn:Kapprox_bi_positiveApproximation}) and (\ref{eqn:Kapprox_xi_positiveApproximation}), 
is negligible in Eq. (\ref{eqn:CMRF_trained_approx}).

The procedure of our method is summarized as follows. 
First, given $\mcal{D}$ we compute $\hat{\bm{c}}$ and $\hat{\bm{d}}$ in Eqs. (\ref{eqn:MaximumCondition_c}) and (\ref{eqn:MaximumCondition_d}). 
Then, using $\hat{\bm{c}}$ and $\hat{\bm{d}}$, 
we compute $\tilde{b}_i^{(K)}(x_i)$ and $\tilde{\xi}_{\{i,j\}}^{(K)}(\bm{x}_{\{i,j\}})$ in Eqs. (\ref{eqn:Kapprox_bi_positiveApproximation}) and (\ref{eqn:Kapprox_xi_positiveApproximation}),
and then $ \tilde{\Psi}_K^{\ddagger}(\bm{x})$ in Eq. (\ref{eqn:Hamiltonian_belief_approx}) for certain $K >0$ and $\varepsilon > 0$. 
Finally, we obtain the CMRF determined in our method by Eq. (\ref{eqn:CMRF_trained_approx}), 
and regard the CMRF as the solution to the inverse problem. 

If one wants to obtain coefficients $\hat{\bm{H}}$ and $\hat{\bm{J}}$, 
one can approximately obtain them by using Eqs. (\ref{eqn:etremal_VariationalBetheFreeEnergy_c}) and (\ref{eqn:etremal_VariationalBetheFreeEnergy_d}) 
together with $\tilde{b}_i^{(K)}(x_i)$ and $\tilde{\xi}_{\{i,j\}}^{(K)}(\bm{x}_{\{i,j\}})$ instead of $\hat{b}_i(x_i)$ and $\hat{\xi}_{\{i,j\}}(\bm{x}_{\{i,j\}})$.

\subsection{Numerical Experiment}
\label{sec:NumricalExperiment}

Let us consider CMRF $P_{\mrm{gen}}(\bm{x}) \propto \exp(-\Psi_{\mrm{gen}}(\bm{x}))$ on an undirected graph $G(V, E)$ with $n = 9$, 
where the energy function is defined as
\begin{align}
\Psi_{\mrm{gen}}(\bm{x}) := -\sum_{i \in V}(x_i - \mu)^2 - \sum_{\{i,j\} \in E}|x_i - x_j|,  
\label{eqh:Hamiltonian_generative}
\end{align}
$\mu = (\beta - \alpha) / 2$, and $\mcal{X} =[\alpha,\beta]= [0,1]$.
Suppose that the CMRF is the generative model lying behind the data points in our numerical experiments. 
We generate $N$ data points, $\mcal{D}$, from the generative model by the Markov chain Monte Carlo method, 
and then solve the inverse problem by the method proposed in the previous section using $\mcal{D}$. 
In the following experiments, we supposed that the CMRF used in solving the inverse problem has the same graph structure as the generative CMRF 
and we used Eq. (\ref{eqn:CosineSystem}) as the orthonormal function system in our method.

\begin{figure}[htb]
\begin{center}
\includegraphics[height=5.0cm]{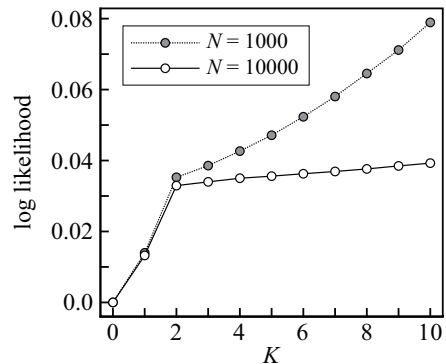}
\end{center}
\caption{Log-likelihood in Eq. (\ref{eqn:EmpiricalError}) versus $K$ on a 1D chain graph.
The plot shows the average over 100 trials.}
\label{fig:likelihood-chain}
\end{figure}

In the first experiment, we supposed that the generative CMRF is defined on a 1D chain graph. 
Because Bethe approximation gives exact solutions in systems with no loops, 
our method described in Sec. \ref{sec:solution_BetheMLE} provides the true solution to the maximum likelihood estimation for the true log-likelihood in Eq. (\ref{eqn:log-likelihood}). 
Given $\mcal{D}$, we computed $\tilde{P}_K(\bm{x})$ in Eq. (\ref{eqn:CMRF_trained_approx}) for a certain $K > 0$ and $\varepsilon = 0.0001$ 
by following the procedure described in Sec. \ref{sec:Approx_For_Implementation}. 
Fig. \ref{fig:likelihood-chain} shows the log-likelihood defined by 
\begin{align}
\tilde{l}_K:= \frac{1}{n N} \sum_{\mu = 1}^N \ln \tilde{P}_K(\mathbf{x}^{(\mu)}),
\label{eqn:EmpiricalError} 
\end{align}
against various $K$. In the computation of $\tilde{l}_K$, we approximately evaluated the partition function in $\tilde{P}_K(\bm{x})$ by the Monte Carlo integration: 
\begin{align}
\int_{\mcal{X}^n}\exp\big( - \tilde{\Psi}_K(\bm{x})\big)d\bm{x} \approx \frac{1}{M}\sum_{m = 1}^M\exp\big( - \tilde{\Psi}_K(\mathbf{y}^{(m)})\big),
\end{align}
where $\mathbf{y}^{(m)}$ is the $m$-th sampled point drawn from the unique distribution over $\mcal{X}$ and $M = 20000$. 
Note that let $\tilde{P}_K(\mathbf{x}^{(\mu)})$ be the unique distribution, $\tilde{P}_K(\mathbf{x}^{(\mu)}) = 1$, when $K = 0$ in this experiment.
The log-likelihood represents the fitness of the solution, $\tilde{P}_K(\bm{x})$, to $\mcal{D}$. 
A solution that gives a higher value of $\tilde{l}_K$ fits the data set better.
$\tilde{l}_K$ increases with the increase in the value of $K$, as shown in Fig. \ref{fig:likelihood-chain}. 
This is because a larger value of $K$ increases the number of controllable parameters and increases the flexibility of the model.
A more flexible model fits the data set better. 
In the plot in Fig. \ref{fig:likelihood-chain} and the following plots, since the error bars (the standard deviations) are too small to be visible, 
we do not show them.

Is the solution with larger $K$ always better? The answer is no in general. 
The important purpose of the inverse problem is to reconstruct the generative model using the given data set. 
It is known that an over-fit to the data set frequently degrades the quality of the reconstruction, because a finite size data set includes noise. 
We measure the quality of the reconstruction, referred to as the generalization error, by the Kullback-Leibler divergence (KLD) defined as
\begin{align}
\mcal{K}_K := \frac{1}{n}\int_{\mcal{X}^n} P_{\mrm{gen}}(\bm{x}) \ln \frac{P_{\mrm{gen}}(\bm{x})}{\tilde{P}_K(\bm{x})} d\bm{x}.
\label{eqn:KLD}
\end{align}
The solution that gives a smaller value of $\mcal{K}_K$ constitutes a better reconstruction of the generative model.
\begin{figure}[htb]
\begin{center}
\includegraphics[height=5.0cm]{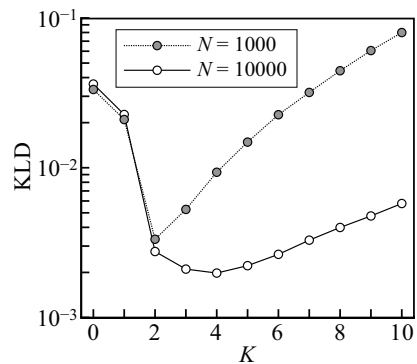}
\end{center}
\caption{Kullback-Leibler divergence in Eq. (\ref{eqn:KLD}) versus $K$ on a 1D chain graph.
The plot shows the average over 100 trials.}
\label{fig:KLD-chain}
\end{figure}
Fig. \ref{fig:KLD-chain} shows the KLD for various values of $K$. 
The KLDs are approximately evaluated by a certain Monte Carlo integration method. 
In the perspective of the generalization error, the optimal value of $K$ is $K = 2$ when $N = 1000$ and $K = 4$ when $N = 10000$.

\begin{figure}[htb]
\begin{center}
\includegraphics[height=5.0cm]{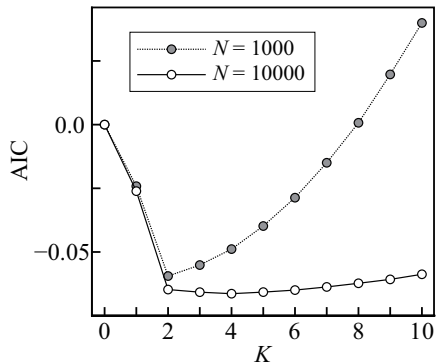}
\end{center}
\caption{Akaike information criterion in Eq. (\ref{eqn:AIC}) versus $K$ on a 1D chain graph.
The plot shows the average over 100 trials.}
\label{fig:AIC-chain}
\end{figure}

In practice, we cannot compute $\mcal{K}_K$, because the generative model is unknown.
The Akaike information criterion (AIC) is one of the most useful criteria of the generalization error~\cite{AIC1973}. 
The AIC is defined as
\begin{align}
\mrm{AIC} := -2  \tilde{l}_K + \frac{2 R_K}{n N},
\label{eqn:AIC}
\end{align}
where $\tilde{l}_K$ is the log-likelihood defined in Eq. (\ref{eqn:EmpiricalError}) and 
\begin{align}
R_K := |\bm{H}| + |\bm{J}| = nK + |E|K^2
\end{align}
is the number of controllable parameters. 
In the context of the AIC, the model that minimizes the AIC is the best in the perspective of the generalization error.
Fig. \ref{fig:AIC-chain} shows the AIC for various $K$. 
We confirm that the AIC is minimized at $K = 2$ when $N = 1000$ and at $K = 4$ when $N = 10000$ 
and that these are consistent with the results in Fig. \ref{fig:KLD-chain}.

\begin{figure}[htb]
\begin{center}
\includegraphics[height=5.0cm]{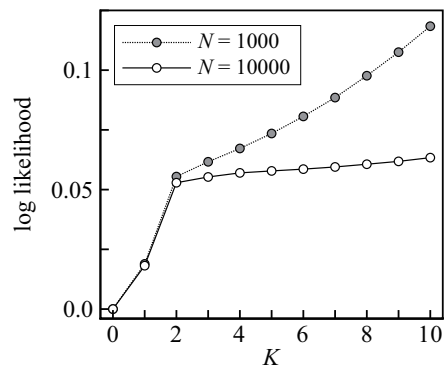}
\end{center}
\caption{Log-likelihood in Eq. (\ref{eqn:EmpiricalError}) versus $K$ on a $3 \times 3$ square grid graph.
The plot shows the average over 100 trials.}
\label{fig:likelihood-grid}
\end{figure}

\begin{figure}[htb]
\begin{center}
\includegraphics[height=5.0cm]{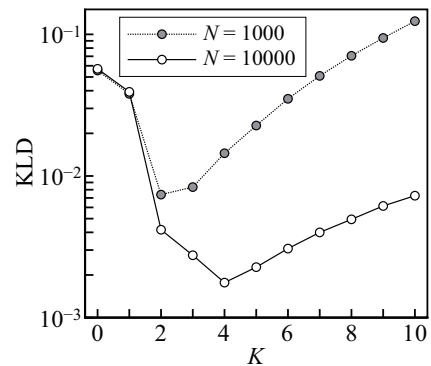}
\end{center}
\caption{Kullback-Leibler divergence in Eq. (\ref{eqn:KLD}) versus $K$ on a $3 \times 3$ square grid graph.
The plot shows the average over 100 trials.}
\label{fig:KLD-grid}
\end{figure}

\begin{figure}[htb]
\begin{center}
\includegraphics[height=5.0cm]{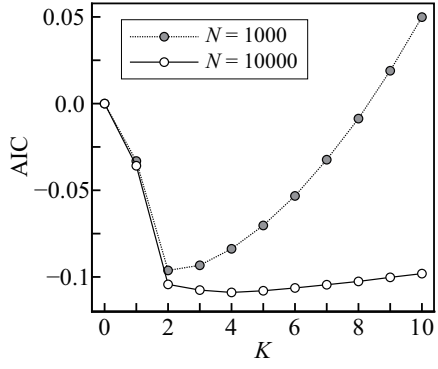}
\end{center}
\caption{Akaike information criterion in Eq. (\ref{eqn:AIC}) versus $K$ on a $3 \times 3$ square grid graph.
The plot shows the average over 100 trials.}
\label{fig:AIC-grid}
\end{figure}

In the next experiment, we supposed the generative CMRF, $P_{\mrm{gen}}(\bm{x})$, is defined on a $3 \times 3$ square grid graph, 
and performed the same numerical experiments as those described above. 
The log-likelihood, the KLD, and the AIC are shown in Figs. \ref{fig:likelihood-grid}--\ref{fig:AIC-grid}, respectively. 
We observe the results similar to those of the first experiment.

\section{Conclusion}
\label{sec:conclusion}

In this paper, we proposed a method for the inverse problem in the CMRF 
with the non-parametrized pair-wise energy function shown in Eq. (\ref{eqh:Hamiltonian}) that uses LBP and orthonormal function expansion. 
As shown in Sec. \ref{sec:solution_BetheMLE}, our method can provide the analytic solution to the inverse problem in the form of an infinite series. 
Since one cannot treat the infinite series computationally, we proposed further approximations, 
the truncation approximation in Eqs. (\ref{eqn:Kapprox_bi}) and (\ref{eqn:Kapprox_xi}) 
and the cut-off approximation in Eqs. (\ref{eqn:Kapprox_bi_positiveApproximation}) and (\ref{eqn:Kapprox_xi_positiveApproximation}), 
for the implementation of our method, in Sec. \ref{sec:Approx_For_Implementation}. 
The numerical results for artificial data were shown in Sec. \ref{sec:NumricalExperiment}. 
From the numerical results, we observed that the optimal value of truncation order $K$ could be found by the AIC. 

However, our method still has a strong limitation, that is, the sample space of the variables is a finite space, $\mcal{X} =[\alpha, \beta]$. 
This limitation was required to impose the normalization constraints and the marginal constraints on the beliefs in Eqs. (\ref{eqn:belief-1_expand}) and (\ref{eqn:belief-2_expand}) for any $\bm{c}$ and $\bm{d}$, 
and this property is quite important for our derivation.
Thus, the extension to the case where $\mcal{X}$ is an infinite space may require an approach different from that used in the current study. 
The extension of our method to such a case will be addressed in our future works.

\appendix

\section{Examples of Orthonormal Function System}
\label{sec:app:Ex_OrthonormalFunctionSystem}

In Sec. \ref{sec:OrthonormalFunctionSystem}, we assume $\mcal{X} = [\alpha , \beta]$ and $\phi_0(x) = 1 / \sqrt{\chi}$. 
One possible choice of the orthonormal function set is the normalized Legendre polynomial, $\phi_s(x) = L_s(x)$, defined as 
\begin{align}
L_s(x):= \frac{1}{2^s}\sqrt{\frac{2s + 1}{2}}\sum_{k=0}^s \binom{s}{k}^2 (x - 1)^{s-k}(x + 1)^k.
\label{eqn:LegendrePolynomial}
\end{align}
The normalized Legendre polynomial satisfies Eq. (\ref{eqn:orthonormalRelation}) on $\mcal{X} = [-1,1]$ and 
it satisfies the recursion formula
\begin{align}
L_{s+1}(x) &= \frac{2s + 1}{s + 1}\sqrt{\frac{2s + 3}{2s + 1}} x L_s(x) \nn
\aleq
- \frac{s}{s + 1}\sqrt{\frac{2s + 3}{2s - 1}}L_{s-1}(x),
\end{align}
for $s \geq 1$, where and $L_0(x) = 1 / \sqrt{2}$ and $L_1(x) = \sqrt{3/ 2}x$.
Another possible choice is $\phi_s(x) = C_s(x)$, where
\begin{align}
C_s(x) := 
\begin{cases}
1 / \sqrt{\pi} & s = 0,\\
\sqrt{2 / \pi} \cos s x & s\geq 1.
\end{cases}
\label{eqn:CosineSystem}
\end{align}
This is the orthonormal function system on $\mcal{X} = [0, \pi]$.

It is noteworthy that, by using linear transformation, we can obtain a new orthonormal function system $\{\tilde{\phi}_s(x)\}$ on $\mcal{X}=[\gamma, \delta]$ 
from an orthonormal function system $\{\phi_s(x)\}$ on $\mcal{X}=[\alpha, \beta]$ as
\begin{align}
\tilde{\phi}_s(x) := \sqrt{\frac{\chi}{\tilde{\chi}}}\phi_s\Big(\frac{\chi}{\tilde{\chi}}x + \frac{\beta \gamma - \alpha \delta}{\tilde{\chi}}\Big),
\label{eqn:orthonormal_translate}
\end{align} 
where $\tilde{\chi}:= \delta - \gamma$. 
The new function system $\{\tilde{\phi}_s(x)\}$ satisfies Eq. (\ref{eqn:orthonormalRelation}) on $\mcal{X} = [\gamma,\delta]$, 
and $\tilde{\phi}_0(x) = 1 / \sqrt{\tilde{\chi}}$, when $\phi_0(x) = 1 / \sqrt{\chi}$.

\section{The Relation in Eq. (\ref{eqn:Hamiltonian_trans2})}
\label{sec:app:RelationHamiltonians}

By using the same technique as in to Eqs. (\ref{eqn:theta_expand}) and (\ref{eqn:wij_expand}), $\ln \hat{b}_i(x_i)$ and $\ln \hat{\xi}_{\{i,j\}}(\bm{x}_{\{i,j\}})$ are expanded as
\begin{align}
\ln \hat{b}_i(x_i) = \sum_{s = 1}^{\infty} A_i^{(s)} \phi_s(x_i) + \mrm{const}
\label{eqn:expand_ln_bi}
\end{align}
and
\begin{align}
\ln \hat{\xi}_{\{i,j\}}(\bm{x}_{\{i,j\}}) &= \frac{1}{\sqrt{\chi}}\sum_{s= 1}^{\infty}\big(B_{\{i,j\}}^{(s,0)} \phi_s(x_i) + B_{\{i,j\}}^{(0,s)}\phi_s(x_j)\big)\nn
\aleq + \sum_{s,t = 1}^{\infty}B_{\{i,j\}}^{(s,t)} \phi_s(x_i)\phi_t(x_j) + \mrm{const},
\label{eqn:expand_ln_xi}
\end{align}
respectively, where $A_i^{(s)}$ and $B_{\{i,j\}}^{(st)}$ are the expanding coefficients defined by
\begin{align*}
A_i^{(s)}&:=\int_{\mcal{X}}\big(\ln \hat{b}_i(x_i)\big) \phi_s(x_i) dx_i, \\
B_{\{i,j\}}^{(s,t)} &:=\int_{\mcal{X}^2} \big(\ln \hat{\xi}_{\{i,j\}}(\bm{x}_{\{i,j\}})\big) \phi_s(x_i)\phi_t(x_j) dx_i dx_j.
\end{align*}
From Eqs. (\ref{eqn:etremal_VariationalBetheFreeEnergy_c}), (\ref{eqn:etremal_VariationalBetheFreeEnergy_d}), and $B_{\{i,j\}}^{(s,t)} = B_{\{j,i\}}^{(t,s)}$, we have
\begin{align}
\hat{H}_i^{(s)} &= (1 - |\partial_i|) A_i^{(s)} + \frac{1}{\sqrt{\chi}}\sum_{j \in \partial_i}B_{\{i,j\}}^{(s,0)},
\label{eqn:relation_H_AB}\\
\hat{J}_{\{i,j\}}^{(s,t)} &= B_{\{i,j\}}^{(s,t)}.
\label{eqn:relation_J_B}
\end{align}
From Eqs. (\ref{eqn:expand_ln_bi})--(\ref{eqn:relation_J_B}), the orthonormal function expansion of the the energy function in Eq. (\ref{eqn:Hamiltonian_belief}) is written as
\begin{align}
\Psi^{\ddagger}(\bm{x})&=-\sum_{i \in V}\sum_{s = 1}^{\infty}\hat{H}_i^{(s)}  \phi_s(x_i) \nn
\aleq 
- \sum_{\{i,j\}\in E} \sum_{s,t = 1}^{\infty} \hat{J}_{\{i,j\}}^{(s,t)} \phi_s(x_i)\phi_t(x_j) + C_1,
\end{align}
where $C_1$ is the constant originates from the constants in Eqs. (\ref{eqn:expand_ln_bi}) and (\ref{eqn:expand_ln_xi}). 
From this equation we obtain Eq. (\ref{eqn:Hamiltonian_trans2}).

\subsection*{Acknowledgment}
This work was partially supported by JST CREST Grant Number JPMJCR1402
and by JSPS KAKENHI Grant Numbers 15K00330, 15H03699, and 15K20870.

\bibliography{citations}

\begin{thebibliography}{20}%
\makeatletter
\providecommand \@ifxundefined [1]{%
 \@ifx{#1\undefined}
}%
\providecommand \@ifnum [1]{%
 \ifnum #1\expandafter \@firstoftwo
 \else \expandafter \@secondoftwo
 \fi
}%
\providecommand \@ifx [1]{%
 \ifx #1\expandafter \@firstoftwo
 \else \expandafter \@secondoftwo
 \fi
}%
\providecommand \natexlab [1]{#1}%
\providecommand \enquote  [1]{``#1''}%
\providecommand \bibnamefont  [1]{#1}%
\providecommand \bibfnamefont [1]{#1}%
\providecommand \citenamefont [1]{#1}%
\providecommand \href@noop [0]{\@secondoftwo}%
\providecommand \href [0]{\begingroup \@sanitize@url \@href}%
\providecommand \@href[1]{\@@startlink{#1}\@@href}%
\providecommand \@@href[1]{\endgroup#1\@@endlink}%
\providecommand \@sanitize@url [0]{\catcode `\\12\catcode `\$12\catcode
  `\&12\catcode `\#12\catcode `\^12\catcode `\_12\catcode `\%12\relax}%
\providecommand \@@startlink[1]{}%
\providecommand \@@endlink[0]{}%
\providecommand \url  [0]{\begingroup\@sanitize@url \@url }%
\providecommand \@url [1]{\endgroup\@href {#1}{\urlprefix }}%
\providecommand \urlprefix  [0]{URL }%
\providecommand \Eprint [0]{\href }%
\providecommand \doibase [0]{http://dx.doi.org/}%
\providecommand \selectlanguage [0]{\@gobble}%
\providecommand \bibinfo  [0]{\@secondoftwo}%
\providecommand \bibfield  [0]{\@secondoftwo}%
\providecommand \translation [1]{[#1]}%
\providecommand \BibitemOpen [0]{}%
\providecommand \bibitemStop [0]{}%
\providecommand \bibitemNoStop [0]{.\EOS\space}%
\providecommand \EOS [0]{\spacefactor3000\relax}%
\providecommand \BibitemShut  [1]{\csname bibitem#1\endcsname}%
\let\auto@bib@innerbib\@empty
%</preamble>
\bibitem [{\citenamefont {Roudi}\ \emph {et~al.}(2009)\citenamefont {Roudi},
  \citenamefont {Aurell},\ and\ \citenamefont {Hertz}}]{Roudi2009}%
  \BibitemOpen
  \bibfield  {author} {\bibinfo {author} {\bibfnamefont {Y.}~\bibnamefont
  {Roudi}}, \bibinfo {author} {\bibfnamefont {E.}~\bibnamefont {Aurell}}, \
  and\ \bibinfo {author} {\bibfnamefont {J.}~\bibnamefont {Hertz}},\
  }\href@noop {} {\bibfield  {journal} {\bibinfo  {journal} {Frontiers in
  Computational Neuroscience}\ }\textbf {\bibinfo {volume} {3}},\ \bibinfo
  {pages} {1} (\bibinfo {year} {2009})}\BibitemShut {NoStop}%
\bibitem [{\citenamefont {Kappen}\ and\ \citenamefont
  {Rodr\'{i}guez}(1998)}]{Kappen1998}%
  \BibitemOpen
  \bibfield  {author} {\bibinfo {author} {\bibfnamefont {H.~J.}\ \bibnamefont
  {Kappen}}\ and\ \bibinfo {author} {\bibfnamefont {F.~B.}\ \bibnamefont
  {Rodr\'{i}guez}},\ }\href@noop {} {\bibfield  {journal} {\bibinfo  {journal}
  {Neural Computation}\ }\textbf {\bibinfo {volume} {10}},\ \bibinfo {pages}
  {1137} (\bibinfo {year} {1998})}\BibitemShut {NoStop}%
\bibitem [{\citenamefont {Parise}\ and\ \citenamefont
  {Welling}(2009)}]{Parise2005}%
  \BibitemOpen
  \bibfield  {author} {\bibinfo {author} {\bibfnamefont {S.}~\bibnamefont
  {Parise}}\ and\ \bibinfo {author} {\bibfnamefont {M.}~\bibnamefont
  {Welling}},\ }\href@noop {} {\bibfield  {journal} {\bibinfo  {journal} {In
  Proc. of the Joint Statistical Meeting 2005 (JSM2005)}\ }\textbf {\bibinfo
  {volume} {4}} (\bibinfo {year} {2009})}\BibitemShut {NoStop}%
\bibitem [{\citenamefont {Yasuda}\ and\ \citenamefont
  {Horiguchi}(2006)}]{Yasuda2006}%
  \BibitemOpen
  \bibfield  {author} {\bibinfo {author} {\bibfnamefont {M.}~\bibnamefont
  {Yasuda}}\ and\ \bibinfo {author} {\bibfnamefont {T.}~\bibnamefont
  {Horiguchi}},\ }\href@noop {} {\bibfield  {journal} {\bibinfo  {journal}
  {Physica A}\ }\textbf {\bibinfo {volume} {368}},\ \bibinfo {pages} {83}
  (\bibinfo {year} {2006})}\BibitemShut {NoStop}%
\bibitem [{\citenamefont {Yasuda}\ and\ \citenamefont
  {Tanaka}(2009)}]{Yasuda2009}%
  \BibitemOpen
  \bibfield  {author} {\bibinfo {author} {\bibfnamefont {M.}~\bibnamefont
  {Yasuda}}\ and\ \bibinfo {author} {\bibfnamefont {K.}~\bibnamefont
  {Tanaka}},\ }\href@noop {} {\bibfield  {journal} {\bibinfo  {journal} {Neural
  Computation}\ }\textbf {\bibinfo {volume} {21}},\ \bibinfo {pages} {3130}
  (\bibinfo {year} {2009})}\BibitemShut {NoStop}%
\bibitem [{\citenamefont {M{\'e}zard}\ and\ \citenamefont
  {Mora}(2009)}]{Mezard2009}%
  \BibitemOpen
  \bibfield  {author} {\bibinfo {author} {\bibfnamefont {M.}~\bibnamefont
  {M{\'e}zard}}\ and\ \bibinfo {author} {\bibfnamefont {T.}~\bibnamefont
  {Mora}},\ }\href@noop {} {\bibfield  {journal} {\bibinfo  {journal} {Journal
  of Physiology-Paris}\ }\textbf {\bibinfo {volume} {103}},\ \bibinfo {pages}
  {107} (\bibinfo {year} {2009})}\BibitemShut {NoStop}%
\bibitem [{\citenamefont {Marinari}\ and\ \citenamefont
  {Kerrebroeck}(2010)}]{Marinari2010}%
  \BibitemOpen
  \bibfield  {author} {\bibinfo {author} {\bibfnamefont {E.}~\bibnamefont
  {Marinari}}\ and\ \bibinfo {author} {\bibfnamefont {V.~V.}\ \bibnamefont
  {Kerrebroeck}},\ }\href@noop {} {\bibfield  {journal} {\bibinfo  {journal}
  {Journal of Statistical Mechanics: Theory and Experiment}\ }\textbf {\bibinfo
  {volume} {2010}},\ \bibinfo {pages} {P02008} (\bibinfo {year}
  {2010})}\BibitemShut {NoStop}%
\bibitem [{\citenamefont {Ricci-Tersenghi}(2012)}]{Federico2012}%
  \BibitemOpen
  \bibfield  {author} {\bibinfo {author} {\bibfnamefont {F.}~\bibnamefont
  {Ricci-Tersenghi}},\ }\href@noop {} {\bibfield  {journal} {\bibinfo
  {journal} {Journal of Statistical Mechanics: Theory and Experiment}\ }\textbf
  {\bibinfo {volume} {2012}},\ \bibinfo {pages} {P08015} (\bibinfo {year}
  {2012})}\BibitemShut {NoStop}%
\bibitem [{\citenamefont {Nguyen}\ and\ \citenamefont {Berg}(2012)}]{Berg2012}%
  \BibitemOpen
  \bibfield  {author} {\bibinfo {author} {\bibfnamefont {H.~C.}\ \bibnamefont
  {Nguyen}}\ and\ \bibinfo {author} {\bibfnamefont {J.}~\bibnamefont {Berg}},\
  }\href@noop {} {\bibfield  {journal} {\bibinfo  {journal} {J. Stat. Mech.:
  Theor. and Exp.}\ }\textbf {\bibinfo {volume} {2012}},\ \bibinfo {pages}
  {P03004} (\bibinfo {year} {2012})}\BibitemShut {NoStop}%
\bibitem [{\citenamefont {Furtlehner}(2013)}]{Cyril2013}%
  \BibitemOpen
  \bibfield  {author} {\bibinfo {author} {\bibfnamefont {C.}~\bibnamefont
  {Furtlehner}},\ }\href {http://stacks.iop.org/1742-5468/2013/i=09/a=P09020}
  {\bibfield  {journal} {\bibinfo  {journal} {J. Stat. Mech.: Theor. and Exp.}\
  }\textbf {\bibinfo {volume} {2013}},\ \bibinfo {pages} {P09020} (\bibinfo
  {year} {2013})}\BibitemShut {NoStop}%
\bibitem [{\citenamefont {Tanaka}(1998)}]{TTanaka1998}%
  \BibitemOpen
  \bibfield  {author} {\bibinfo {author} {\bibfnamefont {T.}~\bibnamefont
  {Tanaka}},\ }\href {\doibase 10.1103/PhysRevE.58.2302} {\bibfield  {journal}
  {\bibinfo  {journal} {Phys. Rev. E}\ }\textbf {\bibinfo {volume} {58}},\
  \bibinfo {pages} {2302} (\bibinfo {year} {1998})}\BibitemShut {NoStop}%
\bibitem [{\citenamefont {Sessak}\ and\ \citenamefont
  {Monasson}(2009)}]{Monasson2009}%
  \BibitemOpen
  \bibfield  {author} {\bibinfo {author} {\bibfnamefont {V.}~\bibnamefont
  {Sessak}}\ and\ \bibinfo {author} {\bibfnamefont {R.}~\bibnamefont
  {Monasson}},\ }\href@noop {} {\bibfield  {journal} {\bibinfo  {journal}
  {Journal of Physics A: Mathematical and Theoretical}\ }\textbf {\bibinfo
  {volume} {42}},\ \bibinfo {pages} {055001} (\bibinfo {year}
  {2009})}\BibitemShut {NoStop}%
\bibitem [{\citenamefont {Yasuda}\ \emph {et~al.}(2012)\citenamefont {Yasuda},
  \citenamefont {Kataoka},\ and\ \citenamefont {Tanaka}}]{Yasuda2012}%
  \BibitemOpen
  \bibfield  {author} {\bibinfo {author} {\bibfnamefont {M.}~\bibnamefont
  {Yasuda}}, \bibinfo {author} {\bibfnamefont {S.}~\bibnamefont {Kataoka}}, \
  and\ \bibinfo {author} {\bibfnamefont {K.}~\bibnamefont {Tanaka}},\
  }\href@noop {} {\bibfield  {journal} {\bibinfo  {journal} {J. Phys. Soc.
  Jpn.}\ }\textbf {\bibinfo {volume} {81}},\ \bibinfo {pages} {044801}
  (\bibinfo {year} {2012})}\BibitemShut {NoStop}%
\bibitem [{\citenamefont {Yedidia}\ \emph {et~al.}(2005)\citenamefont
  {Yedidia}, \citenamefont {Freeman},\ and\ \citenamefont {Weiss}}]{GLBP2005}%
  \BibitemOpen
  \bibfield  {author} {\bibinfo {author} {\bibfnamefont {J.~S.}\ \bibnamefont
  {Yedidia}}, \bibinfo {author} {\bibfnamefont {W.~T.}\ \bibnamefont
  {Freeman}}, \ and\ \bibinfo {author} {\bibfnamefont {Y.}~\bibnamefont
  {Weiss}},\ }\href@noop {} {\bibfield  {journal} {\bibinfo  {journal} {IEEE
  Transaction on Information Theory}\ }\textbf {\bibinfo {volume} {51}},\
  \bibinfo {pages} {2282} (\bibinfo {year} {2005})}\BibitemShut {NoStop}%
\bibitem [{\citenamefont {Pelizzola}(2005)}]{CVM-review2005}%
  \BibitemOpen
  \bibfield  {author} {\bibinfo {author} {\bibfnamefont {A.}~\bibnamefont
  {Pelizzola}},\ }\href {http://stacks.iop.org/0305-4470/38/i=33/a=R01}
  {\bibfield  {journal} {\bibinfo  {journal} {J. Phys. A: Math. and Gen.}\
  }\textbf {\bibinfo {volume} {38}},\ \bibinfo {pages} {R309} (\bibinfo {year}
  {2005})}\BibitemShut {NoStop}%
\bibitem [{\citenamefont {Kikuchi}(1951)}]{CVM1951}%
  \BibitemOpen
  \bibfield  {author} {\bibinfo {author} {\bibfnamefont {R.}~\bibnamefont
  {Kikuchi}},\ }\href {\doibase 10.1103/PhysRev.81.988} {\bibfield  {journal}
  {\bibinfo  {journal} {Phys. Rev.}\ }\textbf {\bibinfo {volume} {81}},\
  \bibinfo {pages} {988} (\bibinfo {year} {1951})}\BibitemShut {NoStop}%
\bibitem [{\citenamefont {Sudderth}\ \emph {et~al.}(2003)\citenamefont
  {Sudderth}, \citenamefont {Ihler}, \citenamefont {Freeman},\ and\
  \citenamefont {Willsky}}]{NonParaBP2003}%
  \BibitemOpen
  \bibfield  {author} {\bibinfo {author} {\bibfnamefont {E.~B.}\ \bibnamefont
  {Sudderth}}, \bibinfo {author} {\bibfnamefont {A.~T.}\ \bibnamefont {Ihler}},
  \bibinfo {author} {\bibfnamefont {W.~T.}\ \bibnamefont {Freeman}}, \ and\
  \bibinfo {author} {\bibfnamefont {A.~S.}\ \bibnamefont {Willsky}},\
  }\href@noop {} {\bibfield  {journal} {\bibinfo  {journal} {Proceedings of the
  IEEE Computer Society Conference on Computer Vision and Pattern Recognition}\
  }\textbf {\bibinfo {volume} {1}},\ \bibinfo {pages} {605} (\bibinfo {year}
  {2003})}\BibitemShut {NoStop}%
\bibitem [{\citenamefont {Ihler}\ and\ \citenamefont
  {McAllester}(2009)}]{PartBP2009}%
  \BibitemOpen
  \bibfield  {author} {\bibinfo {author} {\bibfnamefont {A.}~\bibnamefont
  {Ihler}}\ and\ \bibinfo {author} {\bibfnamefont {D.}~\bibnamefont
  {McAllester}},\ }\href@noop {} {\bibfield  {journal} {\bibinfo  {journal}
  {Proceedings of the 12th International Conference on Artificial Intelligence
  and Statistics}\ ,\ \bibinfo {pages} {256}} (\bibinfo {year}
  {2009})}\BibitemShut {NoStop}%
\bibitem [{\citenamefont {Noorshams}\ and\ \citenamefont
  {Wainwright}(2013)}]{BPforCMRF2013}%
  \BibitemOpen
  \bibfield  {author} {\bibinfo {author} {\bibfnamefont {N.}~\bibnamefont
  {Noorshams}}\ and\ \bibinfo {author} {\bibfnamefont {M.~J.}\ \bibnamefont
  {Wainwright}},\ }\href@noop {} {\bibfield  {journal} {\bibinfo  {journal}
  {Journal of Machine Learning Research}\ }\textbf {\bibinfo {volume} {14}},\
  \bibinfo {pages} {2799} (\bibinfo {year} {2013})}\BibitemShut {NoStop}%
\bibitem [{\citenamefont {Akaike}(1973)}]{AIC1973}%
  \BibitemOpen
  \bibfield  {author} {\bibinfo {author} {\bibfnamefont {H.}~\bibnamefont
  {Akaike}},\ }\href@noop {} {\bibfield  {journal} {\bibinfo  {journal}
  {Proceedings of the 2nd International Symposium on Information Theory}\ ,\
  \bibinfo {pages} {267}} (\bibinfo {year} {1973})}\BibitemShut {NoStop}%
\end{thebibliography}%
\end{document}